\def\year{2020}\relax
\documentclass[letterpaper]{article} 
\usepackage{aaai20} 
\usepackage{times} 
\usepackage{helvet} 
\usepackage{courier} 
\usepackage[hyphens]{url} 
\usepackage{graphicx} 
\usepackage{subcaption}
\usepackage{booktabs}
\usepackage{algorithm2e}
\usepackage{amssymb}
\usepackage{xcolor}
\usepackage{tikz}
\usetikzlibrary{positioning}
\SetKwComment{Comment}{$\triangleright$\ }{}
\usepackage{amsmath}
\usepackage{graphicx} 
\begin{document}
\title{Iterative Data Programming for Expanding Text Classification Corpora}
\author{Neil Mallinar$^\dagger$\thanks{Work done while at IBM Watson}, Abhishek Shah, Tin Kam Ho, Rajendra Ugrani, Ayush Gupta \\
 IBM Watson, New York, NY, $^\dagger$Pryon Research, New York, NY
}
\maketitle

\begin{abstract}
Real-world text classification tasks often require many labeled training examples that are expensive to obtain. 
Recent advancements in machine teaching, specifically the data programming paradigm, facilitate the creation of training data sets quickly via a general framework for building weak models, also known as labeling functions, and denoising them through ensemble learning techniques. 
We present a fast, simple data programming method for augmenting text data sets by generating neighborhood-based weak models with minimal supervision. 
Furthermore, our method employs an iterative procedure to identify sparsely distributed examples from large volumes of unlabeled data. The iterative data programming techniques improve newer weak models as more labeled data is confirmed with human-in-loop. %
We show empirical results on sentence classification tasks, including those from a task of improving intent recognition in conversational agents.
\end{abstract}

\section{Introduction}

A chat bot that automates handling of common customer service requests often uses a text classifier to identify the intent in incoming requests. Many machine learning techniques are suitable for building these classifiers. The biggest bottleneck in applying such classifiers is the availability of labeled training data. 

Labeling vast amounts of text data is expensive, involving manual annotation by domain experts with complex background knowledge. It is even more challenging when the classes are highly skewed, which renders a
sequential process to look for each class largely ineffective, as examples of the minority class are buried in much larger amounts of data of other classes. Deciding between multiple classes for each example, on the other hand, is also stressful as the annotator has to recall the definitions of all classes all the time.

Much work has been done to make the labeling effort easier by expanding the manually provided labels in some automated way. Commonly used data expansion methods include active learning \cite{Karlos2017-activelearning} , self-training \cite{GuzmanCabrera2009-selftraining}, and co-training \cite{co-training}. 
Active learning utilizes a query-selection strategy and an oracle (e.g., a domain expert) to select an important subset of the unlabeled examples to send to the oracle for labeling. The learner is retrained with the newly obtained labels and the process repeats. Without using an oracle, self-training and co-training leverage the more confident decisions of classifiers trained with available examples to provide additional examples to retrain the classifiers. In self-training, an automatic learner is first trained on any available labeled data and evaluated on the unlabeled data. Any unlabeled examples that the learner can classify with confidence passing a threshold are labeled by the decision and used in retraining, and the process repeats. In co-training, multiple classifiers are trained on different views of the labeled data, and the decision with highest confidence on the unlabeled data from the consensus is added to the labeled set.

Semi-supervised learning methods seek to utilize available labels to maximum effect. Examples are techniques like Expectation-Maximization \cite{Nigam2011-emtext} to assist class discovery, and training generative models through data programming \cite{ratner2016-dp,hybrid} with large amounts of unlabeled text data. 
In this paper, we aim to explore the latter, data programming, in an iterative procedure for the purpose of expanding labeled data efficiently.


We design an iterative expansion process that does not utilize the downstream classifier (for which the training data is built) in any way, so as to stay generally applicable to many choices for the classifier, and to avoid computationally expensive training of that classifier over many iterations of expansion.
We experiment with the process using public and private text corpora and confirm that the approach is effective and superior to simpler iterative expansion techniques.


\begin{algorithm*}[t]
 \KwData{Labeled training data $TR$; unlabeled corpus $U$; selection model $S$; labeler $L$; termination criteria $t = 30$; and batch size $b = 10$}
 \KwResult{Augmented labeled training data, $TR'$ }
 
$done =$ \textbf{False} \;

\While{not done}{
  $R = sort_S(U); found=\textbf{False}$ \Comment*[r]{retrieve rankings of elements of $U$}
  \For{$i = 0; i < \text{min}(t, |R|); i = i + b $}{
    $P = label_L(R\big[i:i+b-1\big])$ \Comment*[r]{label small batches from $R$}
    \If{$|P| > 0$ \Comment*[r]{if any positive examples found}}{ 
      $N = R\big[i:i+b-1\big] - P$ \Comment*[r]{get negative examples}
      $TR = TR + P + N$ \Comment*[r]{expand training data}
      $U = U - P - N ; found =$ \textbf{True} \Comment*[r]{remove newly labeled data from $U$}
      \textbf{break} \Comment*[r]{break out of inner for-loop}
    }
  }
  \If{not found}{
    $done =$ \textbf{True} \Comment*[r]{terminate if no positive examples found}
  }
}

\caption{In the expansion procedure, a selection model extracts small, rank-ordered batches of data from an unlabeled corpus that are relevant to the current labeled data. 
These batches are sent, in order, to the labeler for annotation. Once positive labels are found, the selection model is updated with the new data and the procedure is repeated until a termination condition is met. 
Here, termination is when no more positive labels can be found within a maximum number of batches.} 
\label{alg:expand-procedure}
\end{algorithm*}

\section{Data Programming, Weak Supervision, and Machine Teaching}

Previous work has shown that interactive procedures aid creation and bootstrapping of new class-labeled data \cite{williams2015rapidly}. 
They show that domain experts, who are not necessarily knowledgeable about machine learning systems, are able to effectively build classes through a combination of model definition, labeling, active learning, and more. Domain experts are often expensive and their time is valuable. 
Guided learning approaches aim to lessen the amount of data to be labeled, by using search to select relevant examples for labeling, and are shown to be more effective than random labeling and active learning techniques in settings where data exhibits high skew, as ours is given the iterative nature of the classifier building process. \cite{attenberg2010-kdd}. 

In this work, we employ weak supervision techniques to select batches of examples for labeling. Such techniques have recently been reintroduced through the data programming framework \cite{ratner2016-dp,ratner2018-snorkel}, by the notion of ``labeling functions" that generate weak labels. These functions can encode heuristics or can simply be noisy processes for labeling data. Often such heuristics can be given by domain experts, using significantly less of their time. Once many labeling functions are obtained, the next step is to denoise their decisions to obtain a singular label for each example. The work on Snorkel \cite{ratner2018-snorkel} uses a generative model to denoise the ensemble of labeling functions.
After training the generative model, probabilistic labels are obtained per example and are used to train a downstream learner which generalizes the learned heuristics and apply them to any unseen examples. 

These ideas have been used to expand intent training data for conversational agents \cite{Mallinar2019-bootstrapping} and to quickly label large industrial data sets \cite{hybrid}.
We build on these works by using an iterative procedure to automatically construct weak models and show that the Snorkel generative model can select more relevant sentences for labeling than search-based methods alone.

Our work is also connected to the recent machine teaching literature, following definitions and processes provided by \cite{simard2017machine}. 
Machine teaching is defined to be a procedure that can make a ``teacher" more efficient at constructing machine learning models. 
A teacher is defined as somebody who can transfer knowledge (a concept) to a machine learning system (referred to as ``student" or ``learner"). 
Modes of knowledge transfer include selection, where a teacher can find useful examples that exhibit characteristics of a concept, and labeling, where a teacher can provide annotations to directly relate concepts to examples. Earlier definitions of machine teaching are given by \cite{machine-teaching}, where the task is for a teacher that knows the optimal decision boundary between classes in feature space to provide as few examples as possible to the student learner such that the learner can build the same decision boundary. 
Recently, work has been done in iterative machine teaching \cite{iterative-machine-teaching}, showing that employing a continuous loop allows the teacher to react to the student as it learns. 
Summarizing over all of these definitions, the overarching goal of machine teaching is to avoid expensive searches in data space by making use of teacher models that already know the decision boundary, allowing the teacher to show the student as few examples as possible.

The closest existing literature to machine teaching is in the active learning sub-field. 
Active learning operates in a closed-loop where a learning algorithm is trained on any labeled data available. 
That learner then evaluates a large unlabeled corpus and a query-selection strategy is employed to find a small set of examples that will be most useful, if labeled, to add to the training data. 
An oracle, usually a human annotator or crowd-sourced worker, is asked to assign true labels to the selected set of examples. 

We distinguish our work from active learning frameworks as in \cite{settles.tr09-activelearning} by that we never access or query the downstream learner during our procedure. Often such learners require a decent amount of data and tuning on validation sets to be effective models. In our setting, the amount of labeled data provided at the start is very limited ($<20$ examples per class) to the point that even a validation set is impossible to extract.
Effectively, we are asking the question: can a ``teacher" find a ``curriculum" for a concept that is versatile enough for an unknown learner to understand?

\section{Proposed Methods}


We present a procedure for expanding a small set of labeled training data with examples from a larger unlabeled corpus. Our procedure repeats a basic step that consists of leveraging the labeled examples to rank and select the unlabeled examples, and then consulting an oracle for labeling the selected ones. In each step the selection is limited in size to minimize consumption of the oracle resources.
This procedure continues until a termination criteria is reached. The resultant labeled data set is taken to train a downstream text classifier for final evaluation and eventual use in the application.

\subsection{Iterative Expansion Procedure}

We define the main expansion process, provided in Algorithm \ref{alg:expand-procedure}, in the context of binary label data (positive or negative for a specific class). 
For multi-class data, we simply apply the same expansion process independently for each class and merge the resultant positive labeled sets to get our final expanded training set to avoid extreme data skew that might be caused by intermediate steps prior to the completion of expansion of all the classes.

The algorithm takes as input: an initial training set, $TR$, composed of a small number of examples per class; an unlabeled corpus, $U$, to search within for relevant examples; a selection model, $S$, used to find and evaluate relevancy of unlabeled data with respect to labeled data; and a labeler (oracle), $L$, that can provide true labels quickly. The labeler can be expensive to evaluate in practice (often a domain expert), so we include a batch size setting that restrict the per-iteration labeling effort, and a termination criteria to determine when to stop expanding the data.

At the beginning of each iteration we retrieve a rank-ordered set of examples from the unlabeled corpus that are relevant to the current set of labeled training data via the selection model. 
We then iterate, in-order, over these relevant examples in batches of size $b$ which are sent to the labeler for annotation. 
When a batch is found with positive examples we augment the training data set with the newly found positive and negative examples, where the negative set is the complement of the positive set within the batch.

This is repeated until a termination criteria, parameterized by $t$, is met. 
We define $t$ to be the maximum number of consecutive examples we are willing to send to the labeler, in each iteration, without receiving any positive labels back.
We consider the data to be fully expanded when we reach this point.
In this work, we use $t=30$ and $b=10$ to simulate a user looking through three pages of relevant examples with ten examples per page before quitting.

In the case of multi-class data, we convert the data to binary labels per class and use Algorithm \ref{alg:expand-procedure} to expand each of these binarized data sets independently. 
While we track negative expansions relative to each set, we simply take the union of the positive expansions to be the final augmented multi-class training data. 

\begin{figure}
\centering
\captionsetup{labelfont=bf}
\begin{subfigure}{0.39\textwidth}
  \begin{tikzpicture}
    \node (img) {\includegraphics[width=\textwidth]{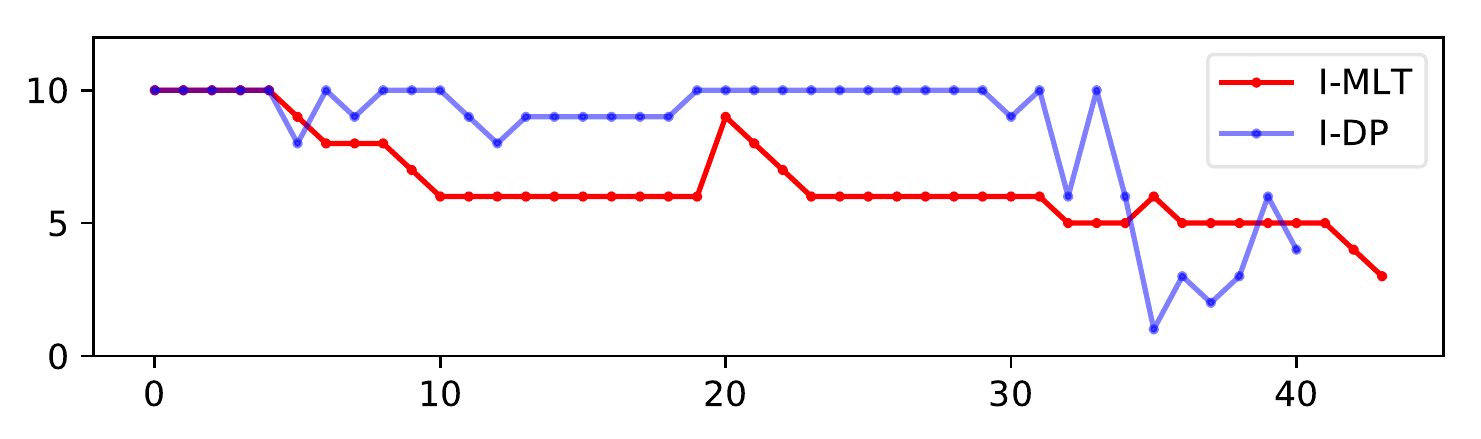}};
    \node[below=of img, node distance=0cm, yshift=1.35cm] {\fontsize{4}{4}\selectfont No. of iterations};
    \node[left=of img, node distance=0cm, rotate=90, anchor=center, yshift=-1.15cm] {\fontsize{4}{4}\selectfont No. of relevant examples};
  \end{tikzpicture}
  \caption{Reddit: graphics\_tablet}
\end{subfigure}
\begin{subfigure}{0.39\textwidth}
  \begin{tikzpicture}
    \node (img) {\includegraphics[width=\textwidth]{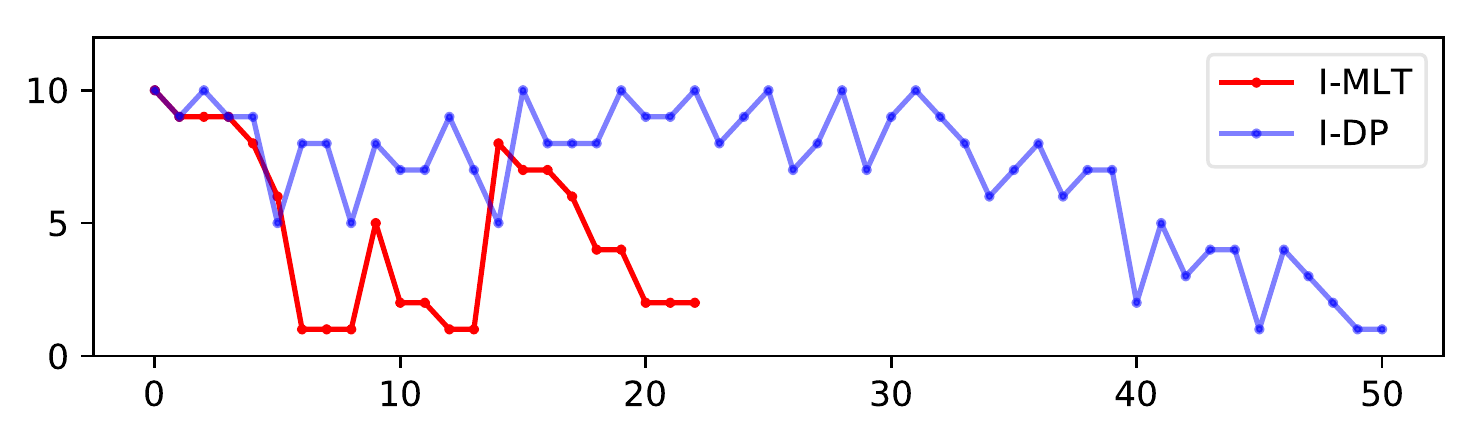}};
    \node[below=of img, node distance=0cm, yshift=1.35cm] {\fontsize{4}{4}\selectfont No. of iterations};
    \node[left=of img, node distance=0cm, rotate=90, anchor=center, yshift=-1.15cm] {\fontsize{4}{4}\selectfont No. of relevant examples};
  \end{tikzpicture}
  \caption{Reddit: home\_security}
\end{subfigure}
\caption{Plotting number of examples added to training data per iteration on Y-axis vs. number of iterations of expansion on X-axis for two classes from the Reddit dataset.}
\label{fig:labeleffort}

\end{figure}
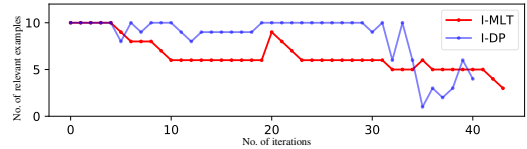
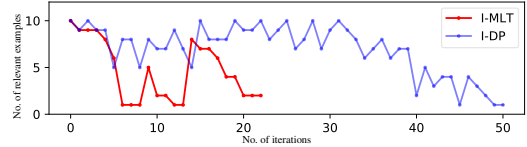

\subsection{Ranking and Selection Methods}

We consider two methods for ranking and selecting the unlabeled examples. The first method uses search-based methods only, while the latter employs data programming.

\subsubsection{Iterative Expansion by Search (I-MLT)}

Search methods have been used to extract data when the class distribution of the unlabeled data is highly skewed \cite{attenberg2010-kdd}.
In expansion by search, we use a search engine as a selection model to find examples relevant to the input examples. 
Here we use the "More Like This" (MLT) feature of the search engine {\it Elasticsearch \footnote{\url{https://www.elastic.co/guide/en/elasticsearch/reference/6.4/search-more-like-this.html}}}
to expand batches of examples and rank them based on the returned score. 
As such, this selection model only uses positively labeled data for expansion and does not make use of negative labels. 

With expansion by search, the unlabeled corpus is ingested into the search engine once and continuously used throughout the process. This allows to perform selection efficiently, compared to re-training a statistical classifier at every iteration.

\subsubsection{Iterative Data Programming with Expansion by Search (I-DP)}

As introduced earlier, data programming proposes to write simple ``labeling functions" that are fast to evaluate and can coarsely assign labels to, potentially large, subsets of unlabeled data. 
We refer to such labeling functions as weak models hereafter, and their label assignment as weak labels (not to be confused with the labels obtained from the labeler $L$ in Algorithm \ref{alg:expand-procedure}, which may also be referred to as strict labels). 
Such weak models generally do not have access to many ground truth labels and make their assignments based on heuristics encoded by human domain experts, crowdsourced workers, or external knowledge bases. We note that while this work uses lexical matching to build neighborhoods, the way we construct weak models can be generalized to any reasonable way to define a subset in a metric space. We leave the exploration of different ways to construct weak models to future work.

Given a collection of $n$ weak models that collectively assign weak labels to $m$ examples, a label matrix, $\Lambda \in \mathbb{R}^{m \times n}$, is constructed out of all of the assignments. 
Where example $i$ is not labeled by weak model $j$, we set $\Lambda_{i, j} = 0$ to represent abstention from labeling the example by the given model.
As such, $\Lambda$ can have varied sparsity depending on how large the coverage of the constituent weak models is. 
As described in \cite{ratner2018-snorkel}, when $\Lambda$ is very sparse or very dense a simple majority vote can be used to denoise the weak labels assigned between the various weak models. 

In the medium density setting, a probabilistic generative model $p_{\theta}(\Lambda, Y)$ is trained over $\Lambda$ and the true labels, $Y$, of the examples. 
It is shown that in the medium density setting, such a model can better denoise the assignments between weak models than a simple majority vote.
When $Y$ is not available, it is treated as a latent variable in the model and optimization is done over the negative log marginal likelihood, $-\log \Sigma_{Y} p_{\theta} (\Lambda, Y)$. 
This is done via a procedure that alternates between Gibbs sampling steps and stochastic gradient descent steps. 

Furthermore, dependencies between weak models are learned through a structure learning process whereby $p_{\theta}(\Lambda, Y)$ is treated as a factor graph.
Additional factors are learned and added to the model to represent common dependencies between weak models \cite{bach2017-structure}.

We construct weak models with two independent MLT expansions on the newly labeled positive and negative examples obtained at each iteration. 
Each search expansion brings a set of ranked examples relevant to the input data. 
Here, the ranking scores from the search engine are discarded and each set of results is weakly labeled as positive or negative, depending on whether the expansion was on positive or negative labeled data, respectively. 
Thus, at the $k^{\text{th}}$ iteration in Algorithm \ref{alg:expand-procedure} we are able to construct $2k$ weak models (half of which are positive expansions, and the other half negative). 
This automatic construction of weak models eliminates the need for domain experts to encode relevant heuristics about the classes, or the need for many crowdsourced workers. 
Instead, a single labeler can quickly label a small number of examples over a few iterations and generate a larger label matrix, $\Lambda$. 

We further truncate the expanded ranked examples list in each weak model to the top $50$ to ensure we retrieve high precision subsets from the unlabeled data and to control the level of noise introduced in the weak labeling process.
After training the generative model, we finally retrieve a list of computed probabilistic labels over examples in our unlabeled corpus (where we default to a score of 0 for those examples that are not covered by at least one weak model). These probabilistic labels are used to rank examples for selection in future iterations of Algorithm ~\ref{alg:expand-procedure}.

We employ this in our procedure to create a hybrid expansion method where rankings obtained both by search expansions and data programming expansions at each iteration are merged into a final ranked examples lists via round-robin merging.
This is done because training the probabilistic model incurs time and computational overhead while using search does not, so we introduce a re-train frequency parameter, $f$, to control how often re-training of the probabilistic model occurs. 

The most recent ranked examples obtained from the generative model are merged with the latest search expansion rankings until the next time the probabilistic model is re-trained (e.g. in the $f$ subsequent iterations of the algorithm after the generative model is trained).
This merging scheme ensures that at each iteration there is an expansion method that reflects the latest qualities of the labeled data (search), as well as an expansion method that produces stronger, less noisy rankings at the cost of being slightly outdated relative to the current set of labeled data (data programming).

The effect of using $f>1$ is two-fold, in that an iteration may not provide enough newly labeled data to be worth retraining the generative model on. 
As such, we continue to expand via search until we have retrieved enough examples to affect the rankings list in a more substantial way. 
In this work, we set $f=3$, meaning the model is only re-trained when there are at least $6$ new weak models to evaluate.


\begin{figure*}[t]
  \captionsetup{labelfont=bf}
  \begin{subfigure}{0.24\textwidth}
    \centering
    \captionsetup{justification=centering}
    \includegraphics[width=\textwidth,height=\textheight,keepaspectratio]{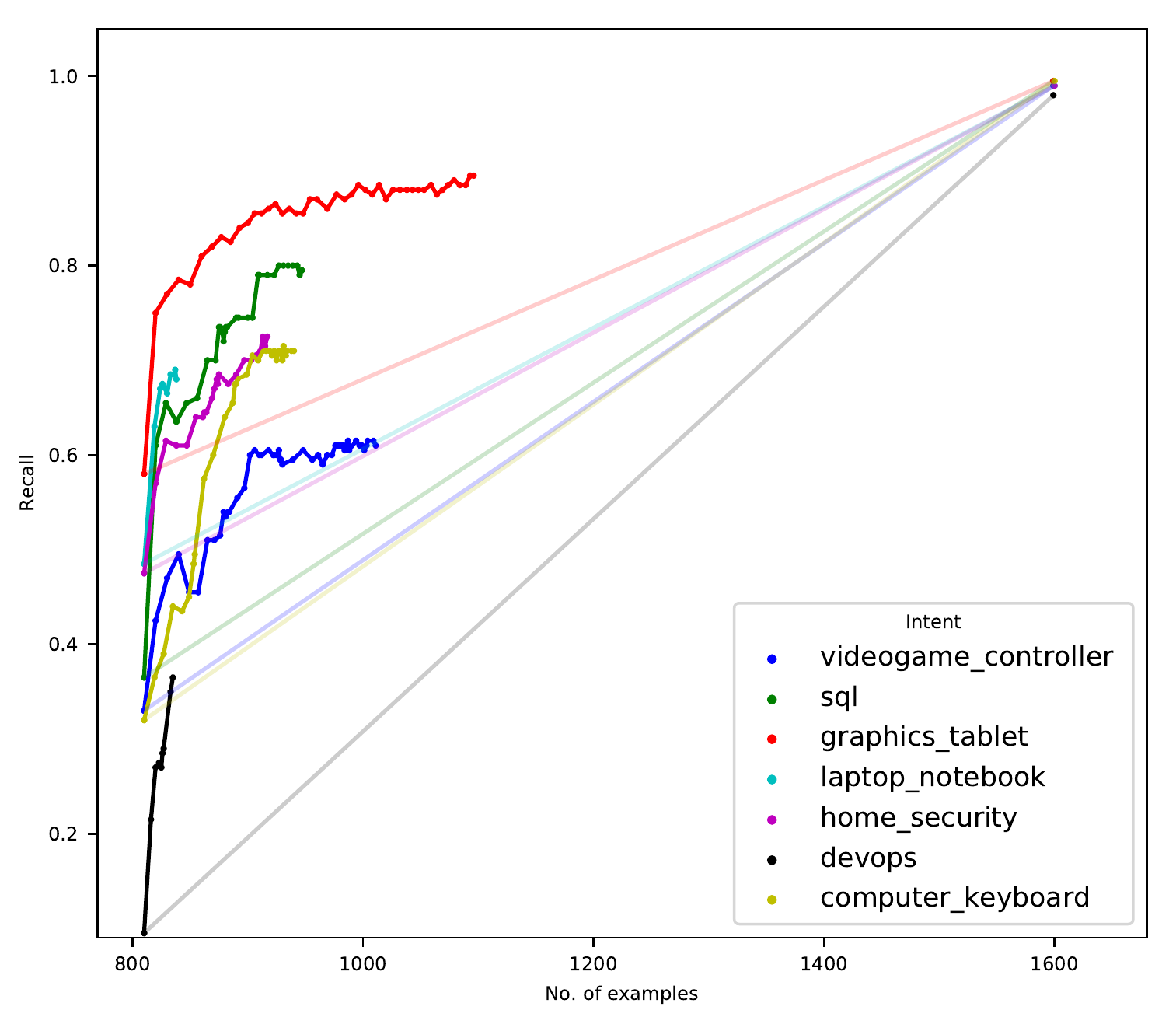}
    \caption{Reddit I-MLT\\Recall vs. \# Examples}
    \label{fig:recall_reddit_sse}
  \end{subfigure}%
  ~ 
  \begin{subfigure}{0.24\textwidth}
    \centering
    \captionsetup{justification=centering}
    \includegraphics[width=\textwidth,height=\textheight,keepaspectratio]{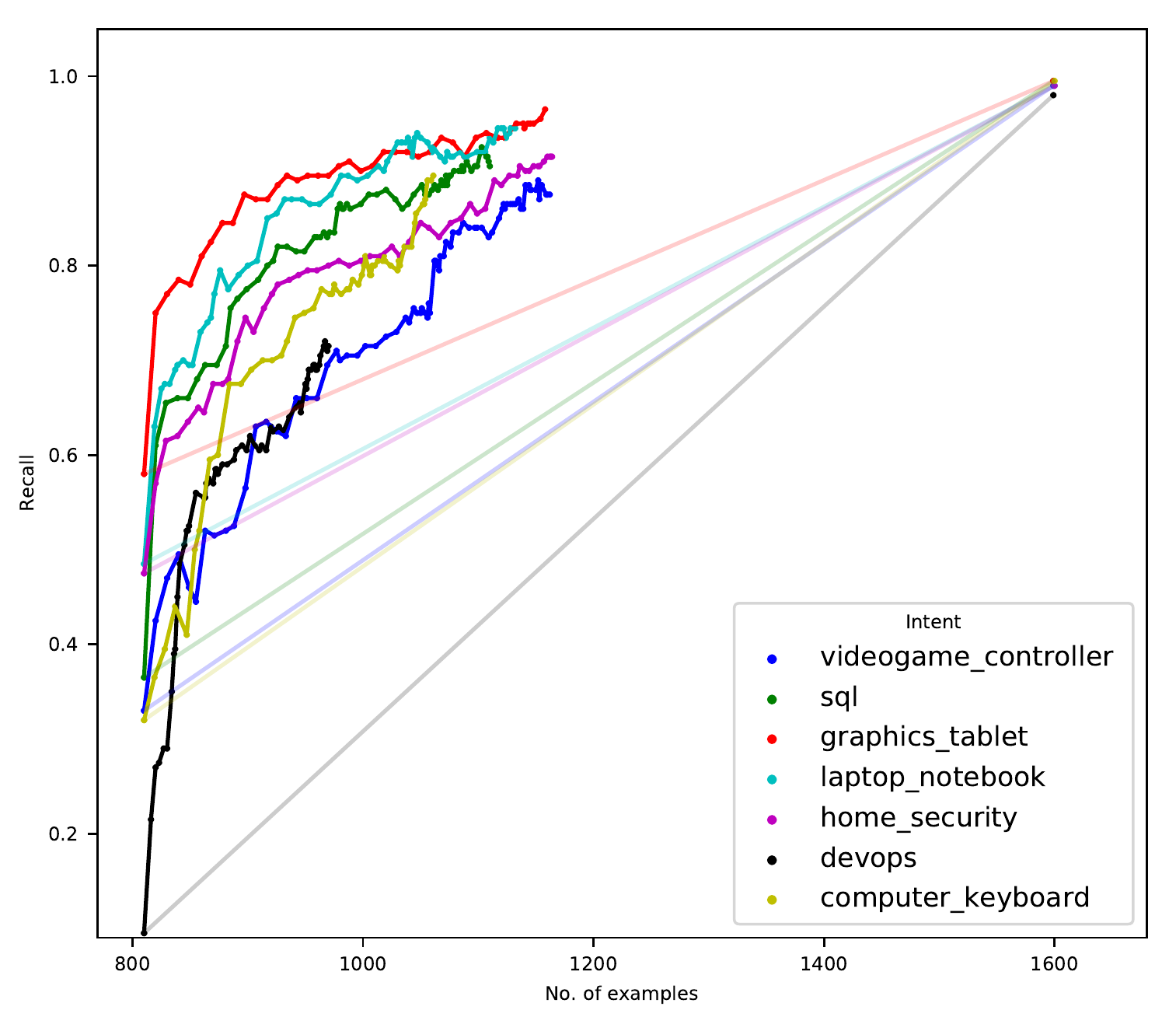}
    \caption{Reddit I-DP \\ Recall vs. \# Examples}
    \label{fig:recall_reddit_wl}
  \end{subfigure}%
  ~ 
  \begin{subfigure}{0.24\textwidth}
    \centering
    \captionsetup{justification=centering}
    \includegraphics[width=0.96\textwidth,height=\textheight,keepaspectratio]{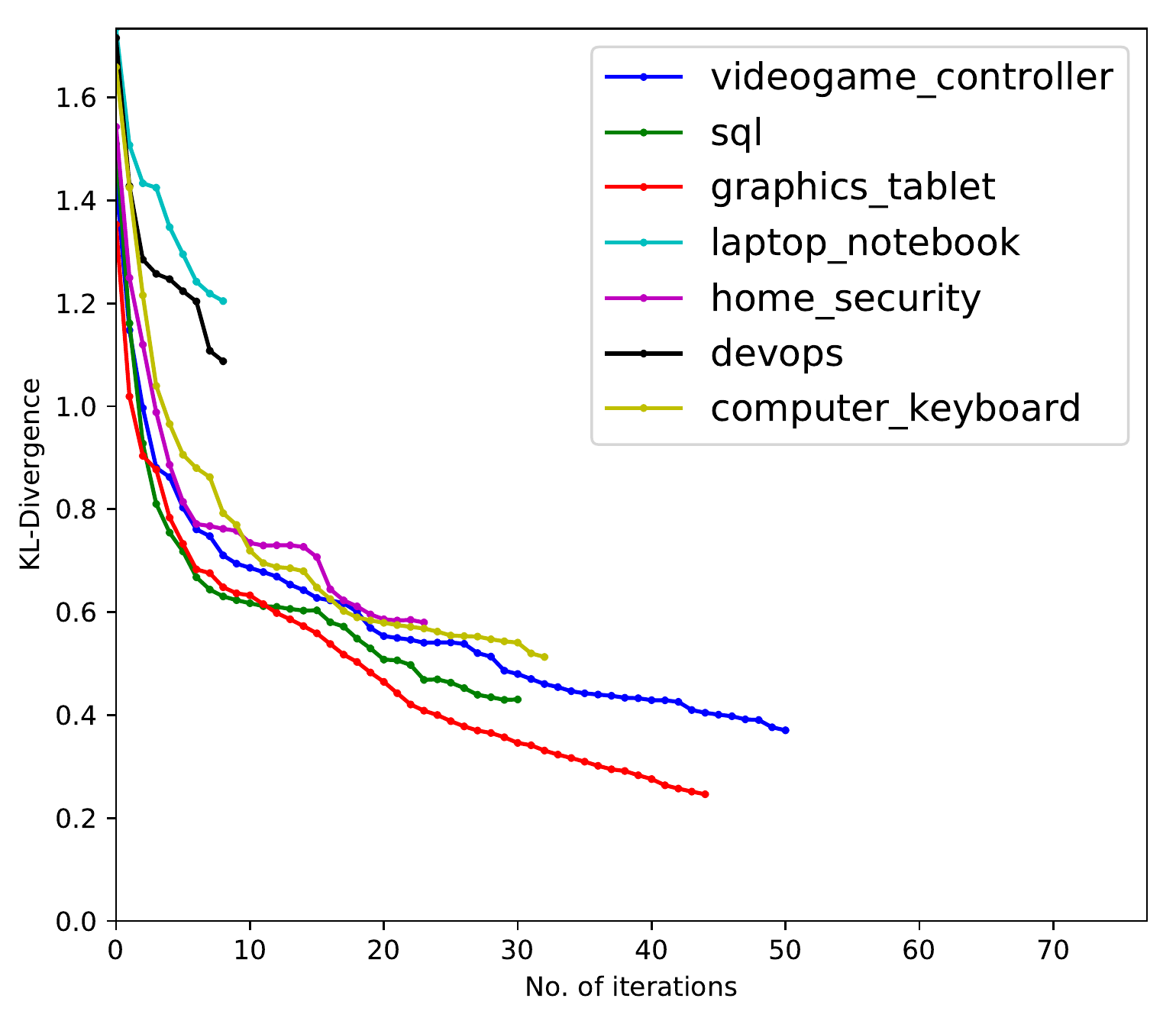}
    \caption{Reddit I-MLT \\ KL vs. \# Iterations}
    \label{fig:kl_reddit_sse}  
  \end{subfigure}%
  ~
  \begin{subfigure}{0.24\textwidth}
    \centering
    \captionsetup{justification=centering}
    \includegraphics[width=0.96\textwidth,height=\textheight,keepaspectratio]{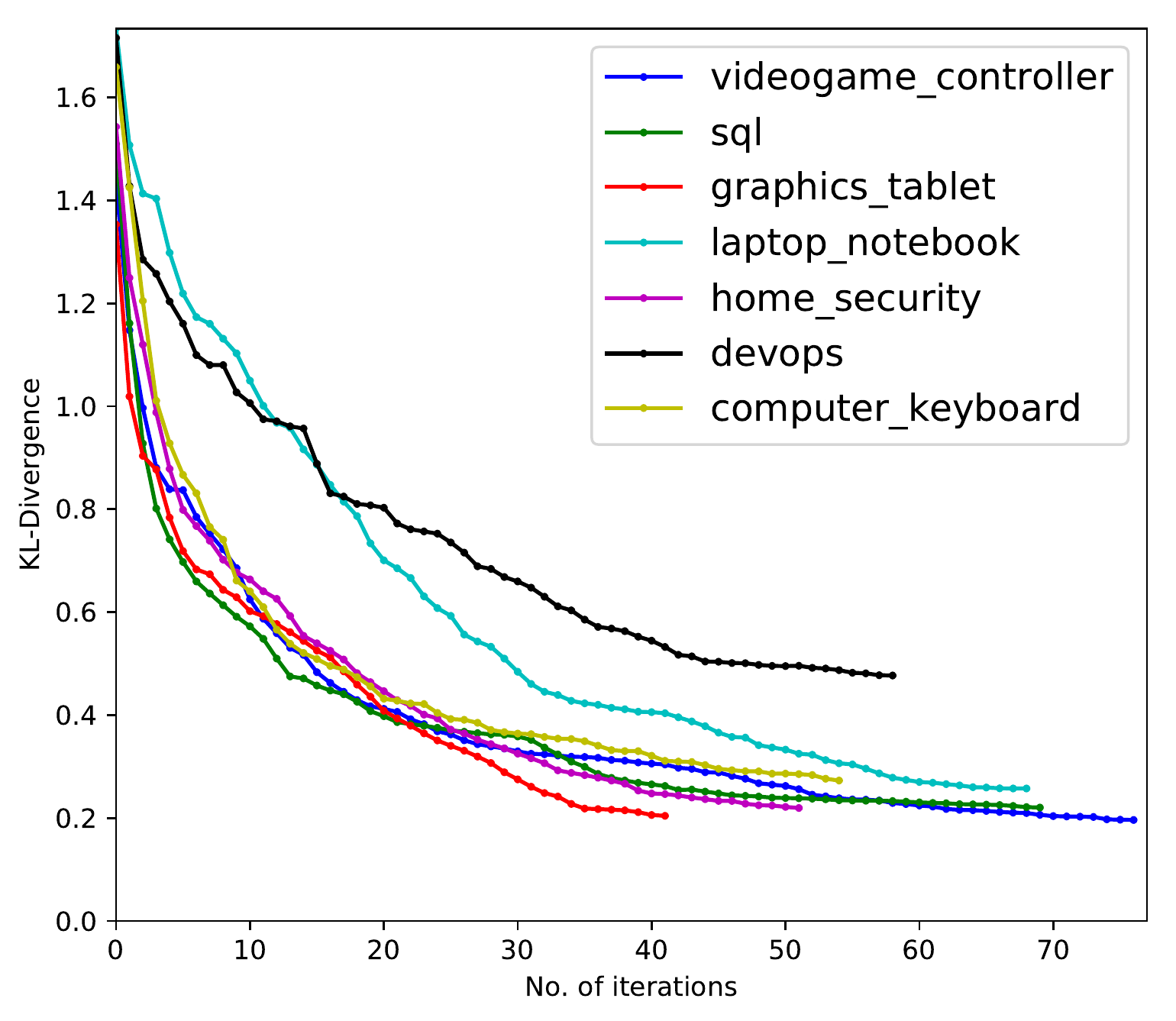}
    \caption{Reddit I-DP \\ KL vs. \# Iterations}
    \label{fig:kl_reddit_wl}  
  \end{subfigure}%
  \caption{Comparison plots for 7 classes sampled from the Reddit data set. \textit{(a,b):} Expansion paths plotting recall on test set vs. number of examples added. The points at the far right of each curve mark the maximum recall if all relevant training samples are found and included. \textit{(c,d):} KL-Divergence vs. number of iterations of expansion. We measure KL-Divergence between the vocabulary distribution at each iteration to the optimal distribution per-class.}
  \label{fig:recall_kl}
\end{figure*}

\section{Experiments and Results}

We use three sentence classification sets for experiments.
From each dataset, we keep only the classes with over 110 examples, and merge the rest into an ``others" class. 
Our experience indicates that most users can afford to hand-write or hand-select a very small set of examples per class. So we hand-select such examples as the initial training set in our experiments. The remaining training examples are stripped of labels and used as the unlabeled corpus ingested into the search engine. 




\textbf{Reddit:} Reddit self-post classification dataset \footnote{The dataset can be downloaded from \url{https://www.kaggle.com/mswarbrickjones/reddit-selfposts/}} has 1,013 sub-reddits from which we choose titles as examples from 81 sub-reddits related to the technology domain. 
We pre-process the titles (and subreddit names as labels) by removing newlines, tabs, special characters, and extraneous spaces and any duplicates.
The final Reddit dataset consists of 810 labeled titles to start with, 63,990 unlabeled titles and 16,200 titles as test set. 


\textbf{Customer Service Inquiries:} We construct two data sets of utterances, one in English and one in Japanese, from privately sourced customer service dialogues. They cover topics from general customer service requests to domain-specific questions in several industries like banking, insurance, and utilities. This data primarily consists of short questions and demands mapped to intents (categories) defined by subject-matter experts. 
The English data set, CS-En, is split into an unlabeled corpus of 15,236 examples, initially labeled set of size 1,998, and test set of size 15,177 spanning 97 categories. 
The Japanese data set, CS-Ja, is split into an unlabeled corpus of 1,709 examples, initially labeled set of size 733, and test set of size 1,691 covering 38 categories that are a subset of those in English. 

\subsection{Experimental Procedure}

To evaluate the effectiveness of our expansion procedure under different selection models, we start with the initial labeled set and expand each class independently as described in Algorithm \ref{alg:expand-procedure}.
After the expansion procedure finishes for each class, we combine all of the fully expanded classes together to create the final labeled training data set. 

Finally, we train a text classifier \footnote{https://www.ibm.com/cloud/watson-assistant/} on this labeled set and evaluate accuracy on the held-out test set.
This accuracy is called the terminal accuracy for the given method used in expansion.
We compare the terminal accuracy to the optimal accuracy, which is obtained by training the same classifier on using the ground truth for the entire corpus from which the examples are selected, and evaluating on the test set.
For further analysis, we measure the KL-Divergence between the vocabulary distribution (at each iteration) of the class being expanded to the optimal vocabulary distribution of that class (as computed from all examples of that class from the corpus).

\def\arraystretch{1.25}
\begin{table}[h]
\centering
\begin{tabular}{llll}
\hline
Methods    & Reddit & CS-En & CS-Ja \\  \hline \hline
Initial    & 0.3035 & 0.7434 & 0.5523
\\
Optimal   & 0.6219 & 0.8725 & 0.7776 \\ \hline
Random    & 0.3070 & 0.7541 & 0.6375 
\\
I-MLT   & 0.4081 & 0.7806 & 0.6606 
\\
I-DP  & \textbf{0.4660} & \textbf{0.8081} & \textbf{0.6972} \\ \hline
\end{tabular}
\caption{Accuracy results for Random, I-MLT, and I-DP on Reddit, English and Japanese Customer Service test sets. We compare the initial and optimal train sets to the terminal train sets, using a downstream classifier for evaluation. \\}
\label{tbl:main-results}

\begin{tabular}{llll}
\hline
Method   & Reddit & CS-En & CS-Ja \\ \hline \hline
Initial  & 810  & 1,998 & 733 \\ 
Optimal  & 64,715 & 17,234 & 2,442 \\ \hline
Random   & 842  & 2,030 & 974 \\
I-MLT  & 6,934 & 6,975 & 1,517 \\
I-DP & 16,006 & 10,125 & 1,769 \\ \hline
\end{tabular}
\caption{Constructed training data size comparisons on our benchmark data for each expansion method. We also present the counts for the initial labeled data set and optimal set.}
\label{tbl:wksp-size-results}
\end{table}

\subsection{Results and Discussion}

We present our main results for I-MLT and I-DP in Table \ref{tbl:main-results}, along with comparisons to a baseline system that uses a random ranker as selection ``model". The accuracy results there reflect the quality of a classifier trained on labeled examples obtained by each method, and are evaluated on the held out examples. Another assessment of the quality of the obtained training set is given by the estimated KL divergence of the word distributions in the training set and the word distribution in the optimal set. 
It is clear in all cases that the I-DP approach is able to construct a training set that performs best on the held-out test sets. 

Figure \ref{fig:labeleffort} shows how many examples are able to be labeled at each iteration of I-MLT and I-DP for two classes sampled from the Reddit data. We see that I-DP is able to consistently find more relevant data for labeling and for more iterations before terminating.
With these results,  we built a tool to assist human annotators to find examples in chat logs to build up intent training data. Early users are finding it very helpful.

Figure \ref{fig:recall_kl} shows the expansion paths, per-class, of seven classes sampled from the Reddit data.
In each plot, the terminal point to the far right of the opaque lines depict the optimal recall of each class, being 1.0 as all training data is included. 
Each plotted point in the main expansion lines represents an iteration step of the expansion procedure. 

One way to assess the ability of these methods is to evaluate the recall of the system at different steps. Also, the assessment on how much recall the end system gives gives a better indication of how much useful data can be pulled from the unlabeled corpus when the data is skewed. Hence, 
we compare the expansion paths of I-MLT in Figure \ref{fig:recall_reddit_sse} to that of I-DP in Figure \ref{fig:recall_reddit_wl} to see how the different selection models affect the rate of recall increase. 
It is evident that with the I-DP procedure, we have a more rapid ascent than I-MLT, and are able to continue to find useful examples for many more rounds before hitting the termination condition.
This can be seen more clearly for the class ``videogame\_controller", where I-MLT seems to converge earlier and reach a recall asymptote whereas I-DP, at the same number of examples sent for labeling, continues to expand. 

We further analyze the effectiveness of each expansion method in Figures \ref{fig:kl_reddit_sse} - \ref{fig:kl_reddit_wl} by plotting the KL-Divergence between the vocabulary distribution at each iteration to the optimal vocabulary distribution. We see that the KL-Divergence curves approach 0 at similar rates as their relative recall curves. We see the KL-Divergence seems to taper off for a few iterations and then begins a steeper descent again, indicating that our procedures are able to explore and push into new areas of the data space and not prematurely terminate. We notice such exploration more so in Figure \ref{fig:kl_reddit_wl}, which utilizes the I-DP expansion process.

While the I-DP algorithm expands the training data more than I-MLT, we see in Figure \ref{fig:recall_kl} that the per-class recall curves, though they taper off as more examples are added, still do not reach convergence yet. This indicates that there is room for future research on steering the expansion. For instance, the method assumes that the full scope of a class can be reached through chains of neighbors.  When this is not true with heavily fragmented classes, we need alternative ways to scatter some seeds to every fragment, by randomization or in-depth analysis of the data characteristics. 
The generative model also needs to be made more efficient if we are to use it in every iteration.

\section{Conclusion}
We present iterative applications, of a search-based selection strategy and a data programming strategy employing weak learning, for expanding text classification data that is independent of knowledge of downstream classifiers.Observations from experiments confirm that the iterative data programming approach can sustain a longer expansion path, allowing the downstream learner to reach better accuracy than the simpler approach. Iterative data programming also leads to better quality training sets sooner, and terminate at obtaining non-exhaustive, labeled training sets with which the learner can attain accuracies very close to those allowed by an optimal training set where all examples must be labeled.

\vspace{-1em}

\bibliography{IAAI-MallinarN.175}
\bibliographystyle{aaai}

\end{document}